\documentclass{esslli}

\usepackage{tabularx}
\usepackage{graphicx}
\usepackage{color}
\usepackage{multirow}
\usepackage{amsmath}
\usepackage{url}
\usepackage{alltt}
\usepackage{listings}
\usepackage{fancybox}

\definecolor{gris}{gray}{0.75}

\newcommand{\dl}[1]{\texttt{#1}}
\newcommand{\dd}{{:}}

\begin{document}

\chapter{Using Description Logics for\\ Recognising Textual Entailment}
\chapterauthor{Paul Bedaride}
\chapteraffiliation{Loria -- UHP Nancy 2 -- Talaris}
\chapteremail{paul.bedaride@loria.fr}

\begin{chapterabstract}
The aim of this paper is to show how we can handle the Recognising Textual Entailment (RTE) task by using Description Logics (DLs). To do this, we propose a representation of natural language semantics in DLs inspired by existing representations in first-order logic. But our most significant contribution is the definition of two novel inference tasks: \emph{A-Box saturation} and \emph{subgraph detection} which are crucial for our approach to RTE.
\end{chapterabstract}

\section{Introduction}

Recognising textual entailment (RTE) is performing the following task:
given two texts $T_1$ and $T_2$ in natural language, determine if we
can infer $T_2$ from $T_1$. As an example consider the three following
sentences:

\begin{itemize}
\item[A:] \emph{``Adam has a son who has a son''},\\[-1.8em]
\item[B:] \emph{``Adam has an offspring who has an offspring''},\\[-1.8em]
\item[C:] \emph{``Adam is a grandfather''}
\end{itemize}

\noindent
We can infer B from A because a son is an offspring. We can also say
that B and C are equivalent because a grandfather is a male who has an
offspring who has an offspring and because \emph{Adam} is a male
name. But we cannot infer A from B or C because an offspring can be a
son or a daughter.

As we can see, recognising textual entailments is far from trivial,
involving many issues that are difficult to solve. The main issue is
that natural language is highly expressive. Due to this expressivity,
it is possible to express the same meaning in several ways, as in B
and C. Furthermore modifying, adding or deleting a word in a sentence
can completely change its meaning (e.g., \emph{Adam (dis)likes
Eve}). Another important issue, related to the first, is that there
exists a huge number of synsets (i.e., sets of words with the same
meaning). It is difficult to exactly map the relation among them
(e.g., an offspring is a child) and to represent all background
knowledge needed for detecting textual entailment.  However, as the
RTE task is widely considered to be relevant for such tasks as
Question-Answering, information retrieval, multi-document
summarization and information extraction, the task has received a
great deal of attention in recent years.

Several different approaches to this task have been proposed and some
of them have been compared in the RTE Pascal
challenge~\cite{DaganGM05}. This challenge compares the
different approaches using a corpus of annotated pairs of texts,
usually referred to as T for Text and H for Hypothesis. For each pair,
it is specified whether T entails H or not. One outcome of this
comparison is that symbolic methods perform better than statistical
methods. Symbolic methods --- using techniques and intuitions rooted
in semantics, syntax, logic, etc --- typically have about 75\%
accuracy. Statistical methods --- based on techniques like n-grams,
lexicon, etc --- have about 60\% accuracy. The method that we describe
in this paper is symbolic. It differs from other symbolic methods
because it uses Description Logics. The first important reason to
choose these logics for the RTE task is that they are decidable and
there exists highly optimized reasoners (e.g., RACER~\cite{HaMo01e})
for different inference tasks. Moreover, we can (at least partially)
represent background knowledge and the semantic representation of
sentences in these logics. Other symbolic techniques which have
already been investigated for the RTE task (e.g., lexical alignment to
detect synonyms), could perhaps be integrated into our approach,
improving the performance, but we are not going to discuss these
possibilities here.

\medskip
As Description Logics (DLs) are the core of our approach to textual
entailment, we will start with a very brief introduction to these
formalisms. Description Logics are formal languages for
knowledge representation. They were inspired by Quillian's semantic
network~\cite{Quillian88} and Minsky's frame
semantics~\cite{Minsky74}. DLs classify knowledge in two
parts: the T-Box and the A-Box. The T-Box contains terminological
information which is general (good for representing background
knowledge). The A-Box contains assertions which are specific (good for
representing sentences). Another way to see the division between these
two kinds of information is to regard the T-Box as rules which govern
our world (e.g., laws from physics, chemistry, biology, etc), and the
A-Box as depicting the world's individuals (e.g., a table, a chair, a
man, etc).

Description Logics employ the notions of concept, role and
individual. Concepts are classes of elements and are interpreted as a
subset of a given universe. Roles are links between elements and are
interpreted as binary relations of a given universe. Individuals are
the elements of a given universe.

A knowledge base $\Sigma$ is a pair $\langle T,A
\rangle$. $T$ is the T(erminological)-Box, a finite set of expressions
called General Concept Inclusions (CGI) with shape $C_1 \sqsubseteq
C_2$ where $C_1$,$C_2$ are concepts. The intended meaning of $C_1
\sqsubseteq C_2$ is that the set of individuals in $C_1$ is included
in the set of individuals in $C_2$. $C_1 \doteq C_2$ is a notation for
$C_1 \sqsubseteq C_2$ and $C_2 \sqsubseteq C_1$. Formulas of $T$ are
also called terminological axioms. $A$ is the A(ssertion)-Box, a
finite set of expressions with shape $a \dd C$ or $(a,b) \dd R$ where
$C$ is a concept, $R$ a role and $a$,$b$ two individuals. The first
expression means that the individual $a$ belongs to the set of
individuals satisfying $C$. The second expression means that the
relation $R$ holds between $a$ and $b$. Formulas of $A$ are called
assertions.

In the description logic that we used, which is known as
$\mathcal{ALCI}$~\cite{2003handbook}, we can form
complex concepts from atomics concepts. They can be made up by
negation ($\neg$), conjunction ($\sqcap$) and disjunction ($\sqcup$)
of concepts. Roles can either be atomic, or the inverse ($R^-$) of an
atomic role. We can also use the universal quantifier ($\forall$
\dl{Role}.\dl{CONCEPT}) to form a complex concept which is true for an
individual $i$ if all roles \dl{Role} which have for first argument
$i$, have for second argument an individual for whom \dl{CONCEPT} is
true. The existential counterpart is  defined as in first order logic:
$(\exists\dl{Role}.\dl{CONCEPT}) \equiv
(\neg\forall\dl{Role}.\neg\dl{CONCEPT})$.

Several reasoning tasks can be handled in DLs once we have defined a
knowledge base $\langle T,A \rangle$. For example, \emph{instance
  checking} tests if an individual is an instance of a specified
concept. \emph{Relation checking} tests if there exists a relation
between two individuals. \emph{Knowledge base consistency} tests if
$\langle T,A \rangle$ is consistent. These tasks can be used for
defining more complex tasks such as \emph{query individuals} which
find all instances of a concept.

We will define two novel reasoning tasks to use DLs for RTE. The most
important of these is the \textit{subgraph detection task}, which we
will discuss in detail later; here we'll introduce the simpler
\emph{A-Box saturation} task. This consists of completing A-Box
information according to a given T-Box. Given a knowledge base
$\langle T,A \rangle$, we say that $A'$ is a saturation if for each
individual $\dl{a}$, atomic concept $\dl{C}$ and role $\dl{R}$
appearing in $\langle T,A \rangle$ there is an assertion $\dl{a} \dd
\dl{C}$ in $A'$ if and only if $\langle T,A \rangle \models \dl{a} \dd
\dl{C}$, and an assertion $(\dl{a},\dl{b})\dd\dl{R}$ in $A'$ if and
only if $\langle T,A \rangle \models (\dl{a},\dl{b})\dd\dl{R}$.

For example we can have the following T-Box:
$$
T = \left\{
\begin{array}{rcl}
 \dl{PARENT} &\doteq& \exists\dl{Parent-of}.\dl{SOMEONE}\\
 \dl{GRANDFATHER} &\doteq& \exists\dl{Father-of}.\dl{PARENT}
\end{array}
\right\}$$

\noindent
expressing respectively that \emph{parent} is equivalent to \emph{someone who is
the parent of someone} and \emph{grandfather} is equivalent to \emph{someone who is
the father of someone who is a parent}. We can also represent the
sentence \emph{``Adam is the father of someone who is the parent of
  someone''} by the following assertions:
$$
A = \left\{
\begin{array}{c}
 \dl{a} \dd \dl{ADAM}, \dl{s1} \dd \dl{SOMEONE}, \dl{s2} \dd \dl{SOMEONE}\\
 (\dl{a},\dl{s1}) \dd \dl{Father-of}, (\dl{s1},\dl{s2}) \dd \dl{Parent-of}
\end{array}
\right\}$$

\noindent
By applying the T-Box to the A-Box we can deduce that \dl{s1} is a
\dl{PARENT} thanks to the first rule and that \dl{a} is a
\dl{GRANDFATHER} thanks to the second rule. If we add the two pieces
of information to the A-Box, we obtain a saturated A-Box.

There exist automatic theorem provers for different DLs including
$\mathcal{ALCI}$ the logic we are going to use. They handle
efficiently several reasoning tasks, including instance and relation
checking, concept and knowledge base consistency, and getting all
instances of a concept. They can also perform A-Box saturation, but
the more complex subgraph detection task will require a new algorithm.

\section{Representation of sentences in DLs}

To start with, we will explain how the meaning of a sentence can be
partially represented as a DL formula. We say partially because the
expressivity of DLs is limited and the meaning of a text is complex,
so our representation is an approximation of the actual meaning of the
text. For instance, many syntactic elements such as articles,
quantifiers, and modalities will not be considered in
our approach. The sentence \emph{``The cat eats an apple''}, for
example, will be approximated by \emph{``cat eat apple''}.

During the definition of our representation we should remember that
our final goal is recognising textual entailment, hence we should
struggle to have the same representation for the same meaning whenever
possible. The main idea of our approach is to represent each sentence
by an A-Box, the background knowledge by a T-Box and then to check if
the model of the entailed sentence is a subgraph of the graph of the
entailing sentence.

We now describe our approach step by step. We first discuss how to
represent sentences in DLs. We start by introducing predicate-arguments
dependencies, then we discuss modifiers, and we finish by explaining
adjectives and negation.

\paragraph{Predicates-arguments dependencies.}
Our representation of sentences is based on Davidson's
semantics~\cite{Davidson80} which represents events as individuals. For
example, the sentence \emph{``John loves Mary''} is represented by the
first order formula $love(e) \wedge john(j) \wedge  mary(m) \wedge
agent(e,j) \wedge patient(e,m)$. Here $e$ stands for the event of
\emph{loving}, $j$ stands for the individual named \emph{John}, and
$m$ stands for the individuals named \emph{Mary}. $j$ is the agent
of the event $e$, and $m$ is its patient.

By using Davidson's semantics we only need to use unary or binary
predicates. This fits well with our DLs approach by making a
correspondence between unary predicates and concepts, and binary
predicates and roles. The sentence \emph{``John loves Mary''} is
represented in DL as $\dl{e} \dd \dl{LOVE}$, $\dl{j} \dd \dl{JOHN}$,
$\dl{m} \dd \dl{MARY}$, $(\dl{e},\dl{j}) \dd \dl{Agent}$,
$(\dl{e},\dl{m}) \dd \dl{Patient}$.

We have agreed then on a semantic form, but we don't know which set of
basic concepts and roles we will use for representing the meaning of
words and the relations between words. To define our signature (i.e.,
the set of basic concepts and roles), we use the linguistic database
FrameNet~\cite{baker98berkeley} based on frame semantics. FrameNet is
composed of semantic frames which involve frame elements and which are
evoked by certain lexical units. For instance, the
\emph{Commercial\_transaction} frame describes a common situation
involving a \emph{buyer}, a \emph{seller}, some \emph{goods} and some
\emph{money} and it is evoked by such words as \emph{buy},
\emph{sell}, \emph{pay}, \emph{cost}, \emph{spend}, etc.

To specify the signature which allows us to represent verb semantics,
we link the frame semantics to our representation, and we link frames
to concepts which represent the sense of verbs, and frame elements to
relations which connect verbs to their arguments.

For example, when we want to represent a verb like \emph{sell}, we start by looking up
in FrameNet the corresponding frame. FrameNet tells us that the
concept \emph{sell} is represented by the frame \dl{COMMERCIAL\_TRANSACTION} and
by the thematic relations \dl{Buyer}, \dl{Seller}, \dl{Goods} and
\dl{Money}. Then for the sentence \emph{``Adam buys chocolate in the
  supermarket for 2 euros''}, we have the following representation as
a DL A-Box:
$$
A = \left\{
\begin{array}{c}
\dl{ct} \dd \dl{COMMERCIAL\_TRANSACTION} \\
\dl{a} \dd \dl{ADAM}, \dl{s} \dd \dl{SUPERMARKET}, \dl{c} \dd \dl{CHOCOLATE}, \dl{p} \dd \dl{2\_EUROS} \\
(\dl{ct},\dl{a}) \dd \dl{Buyer}, (\dl{ct},\dl{s}) \dd \dl{Seller}, (\dl{ct},\dl{c}) \dd \dl{Goods}, (\dl{ct},\dl{p}) \dd \dl{Money}
\end{array}
\right\}$$
\vspace*{-.5cm}

\paragraph{Verb modifiers.}
The meaning of a verb can be affected by modifiers (e.g., place, time,
manner, etc). For example in the sentence \emph{``The dog barks
loudly''}, \emph{loudly} affects the meaning of \emph{bark} by adding
to it the fact that the sound produced by the bark is noisy. We must
be able to say that \emph{``The dog barks loudly''} entails that
\emph{``The dog barks''} but not the converse. Verbs may have many
modifiers of the same type, but this is not a problem with
Davidson-style representations. For each modifier we simply conjoin a
concept which represents the modifier sense to the A-Box individual
corresponding to the verb. For example, the sentence \emph{``John
bought a car on Monday 8 may at 5pm''} has the following representation:
$$
A = \left\{
\begin{array}{c}
\dl{ct} \dd \dl{COMMERCIAL\_TRANSACTION} \sqcap \dl{MONDAY} \sqcap \dl{8\_MAY} \sqcap \dl{5PM} \\
\dl{j} \dd \dl{JOHN}, \dl{c} \dd \dl{CAR} \\
(\dl{ct},\dl{j}) \dd \dl{Buyer}, (\dl{ct},\dl{c}) \dd \dl{Goods}
\end{array}
\right\}$$
\vspace*{-.5cm}

\paragraph{Adjectives.}
In Davidson's approach, adjectives are represented as unary predicates applied to the variable which represents the word to
which the adjective is applied. This representation can easily be
used in DLs for adjectives that modify nouns; such adjectives are
essentially treated in the same way that verb modifiers are. But
adjectives can also occur following the copula as in \emph{``The cat
  is big''}. How do we treat them? As our final goal is to recognize
textual entailment, we have to be able to check that \emph{``The
  big cat''} is equivalent to \emph{``The cat is big''}.

The simplest way to recognise textual entailment is to have the same
representation for the same meaning. We will thus represent adjectives
and the verb \emph{to be} in the same way. That is, for adjectives we add
a concept representing the adjective to the event individual which
represents the word to which the adjective is applied. And we consider
the verb \emph{to be} as an isolated verb; we don't create any individual
for it, but we add to the individual representing the subject of the
verb the concept representing the verb's copula. That is, \emph{``the
  big cat''} and \emph{``the cat is big''} will be represented in the
same way, by: $\dl{c} \dd \dl{CAT} \sqcap \dl{BIG}$.

\paragraph{Negation.}
Even though, we have negation in our representation
language, modeling natural language negation is difficult. The
problem is scope. For example, for the sentence \emph{``The dog
  doesn't bark loudly''} there are two possible interpretations. In
the first interpretation, negation takes narrow scope and applies to
\emph{loudly}. In this reading we mean that the dog barks but
that it doesn't bark loudly. In the second interpretation, negation
takes wide scope and applies to \emph{bark loudly}. It means that we
don't know if the dog barks, but if it barks it doesn't do it loudly.

Scope is an ubiquitous phenomenon in natural language. Besides
negation, it also plays a role for quantifiers and verb arguments
(e.g., \emph{``John sees the girl with the telescope''}). We can try
to analyse all possibilities, but this soon leads to an exponential
blowup (e.g., two negations in a sentence can give rise to four different
interpretations, three negation to eight different meanings, and so
on). Moreover, we must have a YES or NO answer for the RTE task, hence
what should we do if the possibilities don't all agree?

Our choice of representation for negation is motivated by our
mechanism for recognising textual entailment. This mechanism is a mix
between logical implication and syntactic similarity. Let's analyse a
concrete example.  We take the following sentences: (A) ``John didn't
buy a fruit'', (B) ``John didn't buy a fruit at midnight'', (C) ``John
didn't buy an apple'', and (D) ``John didn't buy a big fruit'' because
they represent the most common kinds of scope negation.  With a
standard reading of the sentences we are able to detect the following
entailments (and only those): (A)$\Rightarrow$(B), (A)$\Rightarrow$(C)
and (A)$\Rightarrow$(D).

To detect (A)$\Rightarrow$(B), we must have a logical implication
between the negation of the verb (i.e., \emph{``buy''}) and the
negation of the verb and its modifiers (i.e., \emph{``buy at
midnight''}). So we must have a scope for the negation on verb and its
modifier, because otherwise we won't detect $\neg\dl{BUY} \sqsubseteq
\neg (\dl{BUY} \wedge \dl{MIDNIGHT})$.  To detect (A)$\Rightarrow$(C),
we must have a logical implication between the concept \dl{FRUIT} and
the concept \dl{APPLE}. Lexical knowledge will give us the implication
$\dl{APPLE} \sqsubseteq \dl{FRUIT}$, but we need the contraposed form
$\neg\dl{FRUIT} \sqsubseteq \neg\dl{APPLE}$. So we must have the
negation of concepts associated with verb objects to detect this
textual entailment.  Finally, to detect (A)$\Rightarrow$(D), we must
have a logical implication between the negation of the verb arguments
(i.e., \emph{``fruit''}) and the negation of the verb arguments and
their adjectives. This is similar to the first case, so we have a
scope for the negation on the verb arguments and their adjectives
$\neg (\dl{FRUIT} \wedge \dl{BIG})$.

\section{Representing knowledge}
Now that we have seen how to represent sentences in DLs by encoding
them into the A-Box, we will see how we use background knowledge to
detect textual entailments such as \emph{``a cat eats'' $\Rightarrow$
  ``an animal eats''}. The knowledge required to detect this
entailment is lexical knowledge which explains that a cat is an
animal, thus that the \dl{CAT} concept is subsumed by the \dl{ANIMAL}
concept.

\newcommand{\se}[1]{\textsc{#1}}

We use two repositories of lexical knowledge to detect textual
entailment. The first is FrameNet, which we already used to represent
text with the same meaning in the same way. The second is
WordNet~\cite{lin98wordnet}, which records different lexical relations
between synsets, like synonymy, antonymy or hyponymy. To check that T
entails H, we retrieve \se{slt} and \se{slh}, the synsets list of the
words of T and H using WordNet. For each synset \se{st} and \se{sh} of
\se{slt} and \se{slh} we check if there exists a lexical relation
between them. If there exists a synonymy relation between \se{st} and
\se{sh} we add the following CGI to the background knowledge: $\dl{ST}
\doteq \dl{SH}$. If there exists an antonymy relation between \se{st}
and \se{sh} we add the following CGI to the background knowledge:
$\dl{ST} \sqsubseteq \neg \dl{SH}$ and $\dl{SH} \sqsubseteq \neg
\dl{ST}$. And finally for the hyponymy relation we have three
different cases. If \se{st} is an hyponym of \se{sh} then we get the
following CGI: $\dl{SH} \sqsubseteq \dl{ST}$. If \se{sh} is an
hyponym of \se{st} then we get $\dl{ST} \sqsubseteq \dl{SH}$. And
if \se{sh} and \se{st} share an hyponym we get $\dl{ST} \sqsubseteq
\neg \dl{SH}$ and $\dl{SH} \sqsubseteq \neg \dl{ST}$.

For example, to detect the textual entailment \emph{``a cat eats''
  $\Rightarrow$ ``an animal eats''} we check lexical relations between
  senses of \emph{cat} and \emph{animal} using WordNet. We get that
  \emph{cat} is an hyponym of \emph{animal} and we obtain the CGI:
  $\dl{CAT} \sqsubseteq \dl{ANIMAL}$. By applying this CGI to the
  representation of \emph{``a cat eats''} we obtain the following
  saturated A-Box for the sentence \emph{``a cat eats''}:
$$A = \left\{
\dl{i} \dd \dl{INGESTION},  \dl{c} \dd \dl{CAT} \sqcap \dl{ANIMAL}, (\dl{i},\dl{c}) \dd \dl{Ingestor}
\right\}$$

\section{Inference detection - Subgraph detection}

\begin{figure}
   \includegraphics[width=\linewidth]{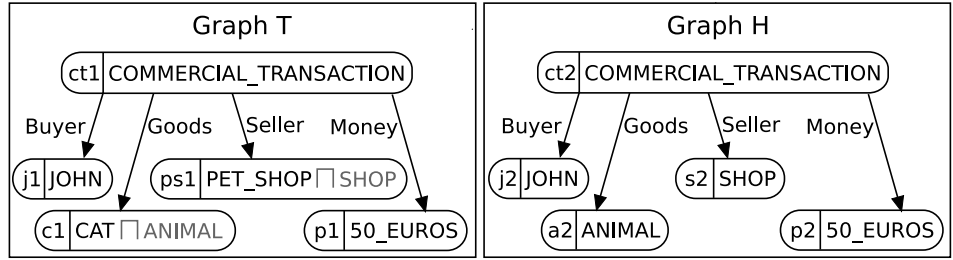}
   \caption{The graph H is a subgraph of the graph T \label{petshop}}
\end{figure}

We have now a way to represent sentences and use background knowledge
to detect textual entailment, and this brings us to the second, and
more complex, of our novel inference tasks: subgraph detection. It
remains to specify how we check if a sentence entails another
sentence. To understand what this involves, we must first note that a
saturated A-Box can be represented as one or more oriented and labeled
graphs (see, for example Figure~\ref{petshop}). What we call subgraph
detection is divided into three steps. First, we create A-Boxes for
the pair (T,H). Then we saturate them with the T-Box created by using
WordNet. Finally we traverse the graphs corresponding to these
saturated A-Boxes to check if the second is a subgraph of the
first. By doing this we verify if all the information in H is also in
T. The algorithm is shown on Figure \ref{algodet}.

We need to do this because existing theorem provers for DLs focus on
tasks which involve one A-Box and one T-Box. There is no existing
tool which handles relations between two DL knowledge bases, and this
is what we required for RTE.

To illustrate our algorithm, we use the example in Figure
\ref{petshop} which aims to show the entailment between the sentence
T:\emph{``John buys a cat at the pet shop for 50 euros''} and the
sentence H:\emph{``A shop sells an animal for 50 euros to
  John''}. These sentences are represented by the following A-Boxes:
$$
T = \left\{
\begin{array}{c}
\dl{ct1} \dd \dl{COMMERCIAL\_TRANSACTION} \\
\dl{j1} \dd \dl{JOHN}, \dl{ps1} \dd \dl{PET\_SHOP}, \dl{c1} \dd \dl{CAT}, \dl{p1} \dd \dl{50\_EUROS} \\
(\dl{ct1},\dl{j1}) \dd \dl{Buyer}, (\dl{ct1},\dl{ps1}) \dd \dl{Seller}, (\dl{ct1},\dl{c1}) \dd \dl{Goods}, (\dl{ct1},\dl{p1}) \dd \dl{Money}
\end{array}
\right\}$$
$$H = \left\{
\begin{array}{c}
\dl{ct2} \dd \dl{COMMERCIAL\_TRANSACTION} \\
\dl{j2} \dd \dl{JOHN}, \dl{s2} \dd \dl{SHOP}, \dl{a2} \dd \dl{ANIMAL}, \dl{p2} \dd \dl{50\_EUROS} \\
(\dl{ct2},\dl{j2}) \dd \dl{Buyer}, (\dl{ct2},\dl{s2}) \dd \dl{Seller}, (\dl{ct2},\dl{a2}) \dd \dl{Goods}, (\dl{ct2},\dl{p2}) \dd \dl{Money}
\end{array}
\right\}$$

\noindent
We compute the background knowledge for detecting the entailment
between these two sentences by using WordNet and we obtain the
following T-Box:
$$
BK = \left\{
\begin{array}{rcl}
 \dl{CAT} &\sqsubseteq& \dl{ANIMAL}\\
 \dl{PET\_SHOP} &\sqsubseteq& \dl{SHOP})
\end{array}
\right\}$$

\noindent
By applying this background knowledge to the DLs representation of the
sentences T and H we obtain the graphs of the Figure \ref{petshop}.

Now we need to check whether a graph \se{gh} is a subgraph of another
graph \se{gt}. Our approach is divided in two parts: node checking and
arc checking.  The first step consists in checking that there exists
a function $f$ which links to each node \se{nh} of \se{gh} a node
\se{nt} of \se{gt} such that the concept associated to \se{nh} is
subsumed by the concept associated to \se{nt}.  The next step is to
check for each arc \se{a} of \se{gh} between nodes \se{n1} and
\se{n2} if there is an arc between $f$(\se{n1}) and $f$(\se{n2})
in \se{gt} which have the same label as \se{a}.
\lstdefinelanguage{algo}
  {morekeywords={if,then,else,elif,def,foreach,do,deftype,return,is},
  sensitive=false,
  morecomment=[l]{//},
  morecomment=[s]{/*}{*/},
  morestring=[b]",
}
\begin{figure}[h!]
\centering

 \begin{tiny}
\doublebox{
\begin{lstlisting}[language=algo]
deftype BIJ = Dict: {IND -> IND}
deftype UNBIJ = Dict: {IND -> (List: IND)}

/* the main function */
def main(t:ABOX, h:ABOX) : BOOL is
  bij:BIJ // nodes that have just one correspondance
  unbij:UNBIJ // nodes that have more than one correspondance
  foreach ind in h.getIndividuals do //get bij and unbij
    inds = t.indsSatisfying(ind.concepts)
    if len(inds)>1 then
      unbij[ind] = inds
    elif len(inds)==1 then
      bij[ind] = inds
    else
      print ``no correspondence for individual '' + ind.name
      stop
  return testAllBijection(bij,unbij)

/* find all bijections and test them */
def testAllBijections (bij:BIJ,unbij:UNBIJ) : BOOL is
  if len(unbij)<=0 then
    testBijection(bij) // we have a bijection and we test it
  else
    ind, List[Ind] = unbij.pop
    foreach i in List[Ind] do
      if testAllBijections(bij+(ind,[i]),unbij) then
        return True
    unbij.append(ind,Lisd[Ind])
    return False

/* test if with this bijection the entailment is correct */
def testBijection(bij:BIJ) : BOOL is
  foreach (src,trg,name) in h.getRelations do: // test all relations
    if not t.hasRelation(bij[src],bij[trg],name) then
      return False
  return True.
\end{lstlisting}
}
\end{tiny}
\caption{Algorithm for subgraph detection \label{algodet}}
\end{figure}

Now that we know how to check if a graph \se{gh} is a subgraph of a
graph \se{gt}, we will check if the graph of the sentence H is a
subgraph of the graph of the sentence T. The first step is to find if
there exists a function $f$. In our example, finding this function is
easy, and it is defined like this: $f(\dl{ct2})=\dl{ct1}$,
$f(\dl{j2})=\dl{j1}$, $f(\dl{a2})=\dl{c1}$, $f(\dl{s2})=\dl{ps1}$,
$f(\dl{p2})=\dl{p1}$. Now for the second step we must check arcs, and
we can see easily that the arcs of the graph of the sentence H exist
in the graph of the sentence T via the function $f$. For instance, the
arc \dl{Buyer} between \dl{ct2} and \dl{j2} exists between
$f(\dl{ct2}$ and $f(\dl{j2})$, that is to say \dl{ct1} and \dl{j1}.

We have used a simple example, but subgraph checking works with more
complex graphs. By more complex graphs we mean graphs containing
identically labelled nodes, or more than one relation between two
nodes. The limit of our algorithm is when we have existentially
quantified information, because in the saturated A-Box we don't expand
existentials. So if we compare the saturated A-Boxes of \emph{``Adam
is the father of someone who is a parent of someone''} and
\emph{``Adam is a grandfather''} we will have many nodes in the first
sentence and only one in the second. Thus the first sentence implies
the second but not the converse.

\section{Tests and Conclusion}
To test our algorithm we have made an implementation in Python which
uses the DL prover RACER. The application takes a file as input which
contains pairs T,H of texts which have been annotated by hand with
respect to whether T entails H or not, and it generates the
semantics\footnote{At present the system doesn't use FrameNet, and
instead we take the verb as concept and basic roles as agent and
patient. Becuase of this we will miss converse cases (i.e., \emph{to sell}
and \emph{to buy}) in our test.} by using the C\&C Tools and Boxer~\cite{cctb}, and adds
the relevant lexical knowledge to detect entailment using
WordNet. After computing all this information, it will use the
algorithm we describe in the previous section to test if T entail H.
We have tested our algorithm on PARC sentence pairs from the
University of Illinois at Urbana-Champaign, which contains 76 pairs
selected to show relevant issues important to the textual entailment.
We can test our implementation on only 75 pairs, as the semantic
generation step fails in one of them. We obtain the following
results:
\begin{center}
 \begin{tabular}{|c|c|c|c|c|}\cline{3-5}
  \multicolumn{2}{c|}{} & \multicolumn{2}{c|}{By Hand} & \multirow{2}{7mm}{Sum} \\\cline{3-4}
  \multicolumn{2}{c|}{}& True & False & \\\hline
  \multirow{2}{20mm}{Application} & True & 23 & 10 & 33 \\\cline{2-5}
   & False & 18 & 24 & 42 \\\hline
  \multicolumn{2}{|c|}{Sum} & 41 & 34 & 75 \\\hline
 \end{tabular}
\end{center}

These tests have shown that what we do works for what we want to do.
That is to say, it works for detecting entailments between simple
sentences (with verbs, noun, verb modifiers, noun modifiers and
negation), with simple lexical knowledge. As we said at the start of
the paper, the present system is not intended to handle entailments
which need complex knowledge, or entailments which hold due to modality,
time expressions, quantification or counting. The incorrect cases in the
test set were usually of this kind.

Our approach is limited by the expressivity of our representation,
which handles only a tiny fragment of the English language. Due to the
expressivity of DLs, some fragments of English will be hard to
represent. For instance, modality needs ideas form modal logic to be
represented correctly (e.g., \emph{``John is an alleged murderer''} is
represented by the formula $john(j) \wedge alleged(murderer(j))$ in
neo-Davison's semantic). Nevertheless, by using more expressives
description logics, we can handle some other fragments, such as
\textit{articulate connective} examples (e.g., \emph{``if Mary comes,
then John comes too''}).

Currently we are working on the implementation of a syntactic analyser
which translates text into our DL representation by using FrameNet,
and testing the use of more expressive logics. For instance, we can
use the one-of operator
$\mathcal{O}$~\cite{2003handbook} to have constraints
on labelled nodes in terminological axioms. This could be useful for
representing sentences with disjunction on individuals like
\emph{``John loves Mary or Jane''}. We can also use hybrid
logics~\cite{areces05:_hybrid_logic} as $\mathcal{H}(@)$ for having
more expressive constraints on labelled nodes, and to represent
articulate connectives.

\bibliographystyle{unsrt} 
\bibliography{biblio}
\nocite{*}

\end{document}